\documentclass[sigconf]{acmart}
\usepackage{color,soul}
\usepackage{tabularx}
\usepackage{afterpage}
\copyrightyear{2026}
\acmYear{2026}

\setcopyright{none}
\acmDOI{10.1145/XXXXXXX.XXXXXXX}
\acmISBN{978-1-4503-XXXX-X/18/06} 
\acmConference[Conference Name]{Conference Long Name}{Date}{Location}
\usepackage{algorithm}
\usepackage{algpseudocode}
\usepackage{booktabs}
\usepackage{graphicx}

\usepackage{float}
\begin{document}

\title{CA-AFP: Cluster-Aware Adaptive Federated Pruning}

\author{Om Govind Jha}
\email{om22@iiserb.ac.in}
\affiliation{
  \institution{IISER Bhopal}
  \city{Bhopal}
  \country{India}
}

\author{Harsh Shukla}
\email{harsh22@iiserb.ac.in}
\affiliation{
  \institution{IISER Bhopal}
  \city{Bhopal}
  \country{India}
}

\author{Haroon R. Lone}
\email{haroon@iiserb.ac.in}
\affiliation{
  \institution{IISER Bhopal}
  \city{Bhopal}
  \country{India}
}

\begin{abstract}
Federated Learning (FL) faces major challenges in real-world deployments due to statistical heterogeneity across clients and system heterogeneity arising from resource-constrained devices. While clustering-based approaches mitigate statistical heterogeneity and pruning techniques improve memory and communication efficiency, these strategies are typically studied in isolation.

We propose CA-AFP, a unified framework that jointly addresses both challenges by performing cluster-specific model pruning. In CA-AFP, clients are first grouped into clusters, and a separate model for each cluster is adaptively pruned during training. The framework introduces two key innovations: (1) a cluster-aware importance scoring mechanism that combines weight magnitude, intra-cluster coherence, and gradient consistency to identify parameters for pruning, and (2) an iterative pruning schedule that progressively removes parameters while enabling model self-healing through weight regrowth.

We evaluate CA-AFP on two widely used human activity recognition benchmarks, UCI HAR and WISDM, under natural user-based federated partitions. Experimental results demonstrate that CA-AFP achieves a favorable balance between predictive accuracy, inter-client fairness, and communication efficiency. Compared to pruning-based baselines, CA-AFP consistently improves accuracy and lower performance disparity across clients with limited fine-tuning, while requiring substantially less communication than dense clustering-based methods. It also shows robustness to different Non-IID levels of data. Finally, ablation studies analyze the impact of clustering, pruning schedules and scoring mechanism offering practical insights into the design of efficient and adaptive FL systems. Our implementation is publicly available at \url{https://anonymous.4open.science/r/CAAFP-C36E/}.
\end{abstract}

\maketitle

%================================================================%
\section{Introduction}
%================================================================%

Federated Learning (FL) has emerged as a crucial paradigm for collaborative machine learning, allowing models to be trained on decentralized devices while preserving user data privacy~\cite{mcmahan2017communication}. This paradigm is particularly well suited for applications that involve sensitive data generated on edge devices such as smartphones, smartwatches, and IoT sensors, which are commonly encountered in domains like Human Activity Recognition (HAR)~\cite{li2023fedchar,kim2021wsn,pandolfo2020analysis}.

The deployment of FL in real-world environments exposes two fundamental challenges: statistical heterogeneity \cite{li2021ditto} and system heterogeneity \cite{wu2023efficient}. Statistical heterogeneity arises when data distributions vary significantly between clients, such as differences in users’ typing patterns or individual activity patterns in HAR applications. System heterogeneity arises due to disparities in hardware and network conditions, including differences in computational speed, memory capacity, bandwidth availability, and energy constraints across devices \cite{bonawitz2019towards}.

Statsitical heterogeneity results in non-IID data which is particularly detrimental to federated optimization thereby slowing convergence and leads to suboptimal global models that fail to generalize well to individual clients~\cite{li2020federated}. To mitigate this issue, Personalized Federated Learning (PFL) approaches have been proposed, which aim to tailor models to individual clients or groups of clients~\cite{li2021ditto}. Under this line of research, clustering-based FL has emerged as a prominent strategy that groups clients with similar data distributions, enabling more relevant and effective aggregation. Representative methods such as FedCHAR demonstrate that clustering clients based on update similarity can significantly improve personalization, robustness under heterogeneous data settings~\cite{li2023hierarchical}.

In real-world FL systems, both statistical heterogeneity and system heterogeneity must be addressed jointly. Recent approaches such as SAFL~\cite{Li2025SAFL} and FLCAP~\cite{miralles2023flcap} attempt to tackle statistical heterogeneity through client clustering and system heterogeneity through model pruning. However, both frameworks suffer from intrinsic limitations in how pruning decisions are made. In particular, pruning is performed while model parameters and cluster assignments are still evolving, causing sparsification to rely on early-stage and potentially unstable optimization signals. SAFL primarily leverages structural cues derived from batch normalization statistics, whereas FLCAP relies mainly on weight magnitude to estimate parameter importance. As a result, these methods fail to capture collective relevance across clients within the same cluster. Consequently, parameters that are weak at the individual client level but consistently important at the cluster level may be prematurely removed. Moreover, the lack of a controlled and gradual sparsification mechanism can lead to instability and degraded accuracy under non-IID data distributions.

To address these limitations, we propose \emph{CA-AFP} (Cluster-Aware Adaptive Federated Pruning), which decouples cluster formation from sparsification and introduces a context-aware pruning strategy. CA-AFP first allows client models to stabilize during an initial dense training phase, after which cluster assignments are formed and pruning is initiated. This design ensures that importance estimates are derived from reliable and converged learning behavior. Furthermore, CA-AFP employs an iterative adaptive masking schedule that dynamically updates sparsity patterns during training, enabling the model to ``heal'' by reactivating previously pruned weights that later become important.

We evaluate CA-AFP on two standard human activity recognition benchmarks, WISDM and UCI-HAR, which represent realistic and controlled federated settings with strong inter-user heterogeneity. We compare against both \textit{dense clustering-based} baselines (FedCHAR~\cite{li2023fedchar}, ClusterFL~\cite{ouyang2023clusterfl}) and \textit{pruning-based} baselines (EfficientFL~\cite{wu2023efficient}, FedSNIP~\cite{fedsnip}), enabling a fair comparison across methods that address statistical and system heterogeneity in isolation. While dense baselines focus on mitigating statistical heterogeneity using fully parameterized models, pruning baselines primarily target system heterogeneity and communication efficiency through sparsification.

Experimental results show that CA-AFP consistently matches or outperforms pruning-based baselines while approaching the accuracy of dense methods at comparable sparsity levels. On both datasets, CA-AFP achieves improved fairness over EfficientFL and FedSNIP while maintaining substantially lower communication costs than dense approaches, even under limited fine-tuning. With additional fine-tuning, CA-AFP further narrows the performance gap with dense baselines. These results demonstrate that CA-AFP effectively preserves parameters that are collectively informative at the cluster level and provides flexible control over the trade-off between accuracy and communication efficiency under device constraints. To support reproducibility, we release the complete implementation of CA-AFP at \url{https://anonymous.4open.science/r/CAAFP-C36E/README.md}.

%================================================================%
\section{Related Work}
%================================================================%

\noindent\paragraph{\bf Cluster-Based Federated Learning:}
Cluster-based federated learning (FL) addresses statistical heterogeneity. For example FedCHAR \cite{li2023fedchar} performs hierarchical clustering based on cosine similarity of model update directions, enabling intra-cluster aggregation that improves robustness by isolating malicious clients while enhancing fairness across participants \cite{li2023hierarchical}. Similarly, ClusterFL \cite{ouyang2021clusterfl} formulates FL as a clustered multi-task optimization problem and jointly learns cluster assignments and model parameters via ADMM, allowing collaborative learning within clusters and discarding slow or weakly correlated clients to reduce communication rounds and improve accuracy \cite{ouyang2023clusterfl}.

Recent clustered FL (CFL) frameworks further refine client grouping by explicitly modeling gradient variance and partial participation. Early CFL approaches assume low intra-cluster gradient variance to achieve faster convergence and improved generalization under non-IID data. Building on this assumption, StoCFL \cite{zeng2025stocfl} introduces a stochastic clustering mechanism that leverages cosine similarity of gradient-based data distribution representations and supports partial client participation. Through a bi-level optimization strategy, StoCFL enables knowledge sharing between cluster-specific models and a global model, improving data efficiency while maintaining cluster specialization \cite{zeng2025stocfl}. Complementarily, FedClust \cite{islam2024fedclust}adopts a one-shot, weight-driven hierarchical clustering strategy that exploits correlations between partial model weights—particularly classifier layers—and underlying client data distributions, achieving stable cluster formation with significantly reduced communication overhead and dynamic support for newly joining clients.

Despite their effectiveness in mitigating statistical heterogeneity and improving convergence under non-IID data, these clustered FL approaches uniformly rely on fully dense model exchanges. They primarily focus on statistical, optimization, and security-related challenges, while overlooking memory footprint and communication efficiency. The absence of sparsification, pruning, or compression mechanisms limits their applicability in resource-constrained federated environments, such as edge devices with strict bandwidth and memory constraints.

\noindent\paragraph{\bf Pruning-based Federated Learning:}
Pruning-based federated learning reduces the communication, computation, and memory overhead of federated optimization by allowing clients to train and transmit sub-models derived from a global model \cite{Huangetal2025,Baietal2025}. A representative approach is Federated Dropout \cite{Cheng2022}, where clients randomly drop neurons or filters during local training and only communicate the remaining parameters, thereby lowering resource requirements while maintaining convergence properties of the global model \cite{caldas2019feddropout}. To further optimize the initial phase of training, FedSNIP \cite{fedsnip}employs single-shot network pruning based on connection sensitivity to establish sparse structures before communication begins. Furthermore, EfficientFL \cite{Li2022EfficientFL} reduces the burden on edge devices by identifying a ``winning subnetwork'' at the start of training, ensuring that only the most impactful parameters are updated and transmitted throughout the federated loop. Such approaches demonstrate that partial model participation can be effective in federated settings, particularly for resource-constrained edge devices.

\noindent\paragraph{\bf Adaptive Federated Pruning:}

Adaptive federated pruning extends pruning-based federated learning by dynamically adjusting model sparsity throughout the training process in response to system constraints, optimization progress, and device heterogeneity \cite{Huang2025FedAPTA}. Unlike static pruning strategies that rely on training dense models followed by one-shot sparsification, adaptive methods integrate pruning directly into the federated optimization loop, enabling devices to progressively learn sparse models without incurring the cost of dense training \cite{huang2024fedmef}. Recent frameworks such as FLCAP \cite{Kapoor2023FLCAP} further refine this by utilizing clustered adaptive pruning to handle scalable systems, while SAFL \cite{Li2025SAFL} employs structure-aware pruning to align sub-models with specific client clusters.

FedResCuE \cite{zhu2022fedrescue} addresses system heterogeneity and unstable network connections by enabling clients to train self-distilled neural networks that can be structurally pruned to arbitrary sizes without requiring fine-tuning. Through progressive learning, predictive knowledge is captured in nested sub-models, allowing clients with different resource budgets to participate using differently sized pruned models while still contributing meaningful updates to the global aggregation process. This adaptive pruning paradigm allows federated learning systems to gracefully accommodate heterogeneous devices and partial model transmissions.

\noindent\paragraph{\bf Cluster-Aware Pruning in Federated Learning:}
Cluster-aware federated learning aims to mitigate statistical heterogeneity by grouping clients with similar data distributions and restricting model aggregation within each cluster \cite{Lu2025FedCPC}. Representative works such as ClusterFL formulate federated learning as a clustered multi-task optimization problem, where the server jointly learns cluster assignments and cluster-specific models using distributed optimization techniques, enabling collaborative learning among similar clients while reducing communication overhead through cluster-wise straggler dropout and node selection \cite{ouyang2021clusterfl}. 

These clustering-based approaches demonstrate that exploiting intrinsic similarity among clients is effective for improving convergence, personalization, and robustness in federated learning.
However, existing cluster-aware federated learning methods primarily operate on fully dense models and focus on statistical heterogeneity, fairness, and security aspects \cite{Zhang2024FedAC}. They do not incorporate pruning or sparsification mechanisms within or across clusters, resulting in substantial memory and communication overhead when deployed on resource-constrained edge devices \cite{ouyang2021clusterfl,li2023fedchar}. Consequently, although clustering reduces harmful interference among heterogeneous clients, the absence of cluster-aware pruning limits the scalability and efficiency of these approaches. This observation highlights a critical research gap in jointly leveraging client clustering and adaptive pruning to achieve both statistical efficiency and system-level resource efficiency in federated learning.

\begin{figure}[t]
    \centering
    \includegraphics[scale=0.37]{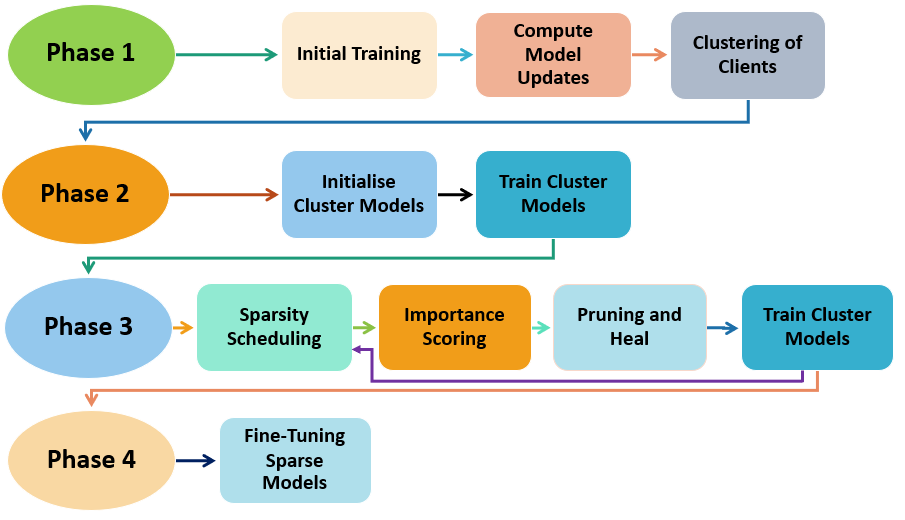}
    \caption{The overall workflow of CA-AFP.}
    \label{fig:method_workflow}
\end{figure}

%================================================================%
\section{Methodology}
%================================================================%
The CA-AFP framework operates in four distinct phases. Algorithm~\ref{alg:caafp_full} (Appendix) provides detailed pseudocode, and Figure~\ref{fig:method_workflow} presents an overview of the workflow.

\subsection{Phase 1 : Initial Training and Clustering}

\textbf{Initial Training}: 
Training proceeds in communication rounds. At each round $t$, the server broadcasts the global model $w^{(t-1)}$ to all the clients. Each client $k$ performs $E$ local training epochs on its dataset $\mathcal{D}_k$ by minimizing the standard cross-entropy loss:

\begin{equation}
\mathcal{L}_k^{\text{CE}}(w_k; \mathcal{D}_k)
= -\frac{1}{|\mathcal{D}_k|}
\sum_{(x,y) \in \mathcal{D}_k} \log p(y \mid x),
\end{equation}

where $\mathcal{L}_k^{\text{CE}}({w}_k; \mathcal{D}_k)$ denotes the cross-entropy loss on client $k$, ${w}_k$ represents the model parameters, $\mathcal{D}_k$ is the local dataset of client $k$ with $|\mathcal{D}_k|$ samples, $(x,y)$ denotes an input--label pair, and $p(y \mid x)$ is the predicted probability of label $y$ given input $x$. 

After completing $E$ local epochs, client $k$ obtains updated parameters $w_k^{(t)}$, which are sent to the server. The server aggregates the client models as

\begin{equation}
w^{(t)} = \sum_{k=1}^K \frac{|\mathcal{D}_k|}{|\mathcal{D}|} w_k^{(t)},
\end{equation}

where $w^{(t)}$ denotes the aggregated global model parameters at communication round $t$, $K$ is the total number of participating clients, and $|\mathcal{D}|=\sum_{k=1}^K |\mathcal{D}_k|$ is the total number of training samples across all clients.
The initial training phase is performed for $T_0$ communication rounds, where $T_0 \geq 0$.

\vspace{0.3cm}

\textbf{Clustering}: 
After the initial training phase of $T_0$ rounds, the final global model $w^{(T_0)}$ is used as the reference model for client clustering. If $T_0=0$, $w^{(T_0)}$ corresponds to a randomly initialized model.

Each client $k$ initializes its model with $w^{(T_0)}$ and performs one local training epoch on $\mathcal{D}_k$, producing updated parameters $w_k^{(T_0, 1)}$. The local update used for clustering is defined as

\begin{equation}
\Delta w_k
=
w_k^{(T_0, 1)} - w^{(T_0)}.
\end{equation}

The pairwise cosine distance between clients $k$ and $k'$ is computed as

\begin{equation}
d(k, k') = 1 -
\frac{
(\Delta w_k)^\top \Delta w_{k'}
}{
\|\Delta w_k\|_2 \,
\|\Delta w_{k'}\|_2
}.
\end{equation}

Here, $\Delta w_k$ and $\Delta w_{k'}$ denote the one-epoch local update vectors of clients $k$ and $k'$, respectively. The operator $(\cdot)^\top$ denotes vector transpose, $(\Delta w_k)^\top \Delta w_{k'}$ represents their inner product, and $\|\cdot\|_2$ denotes the Euclidean ($L_2$) norm.

This distance measures the angular dissimilarity between client updates while being invariant to their magnitudes. Agglomerative hierarchical clustering with average linkage is then applied to the resulting distance matrix to assign each client to a cluster. The clustering process only determines client assignments and does not modify the global model weights.

\subsection{Phase 2: Cluster-Level Stabilization}

After client clustering, CA-AFP initializes one cluster model for each cluster using the Phase~1 global model $w^{(T_0)}$. Each cluster model is then refined through a few rounds of ``dense federated training''~\cite{li2023hierarchical} within its corresponding client group, without any pruning applied at this stage.

Specifically, for each cluster $c$, let $\mathcal{C}_c$ denote the set of clients assigned to cluster $c$. After each round $t$, the server broadcasts the current cluster model $w_c^{(t)}$ to all clients $k \in \mathcal{C}_c$. Within each cluster, each client maintains two models: a personal reference model $w_{p,k}$ and a cluster model $w_c$.

Each client performs local training on its dataset $\mathcal{D}_k$ by minimizing the regularized objective

\begin{equation}
\mathcal{L}_k^{c}(w_c; \mathcal{D}_k)
=
\mathcal{L}_k^{\text{CE}}(w_c; \mathcal{D}_k)
+
\frac{\lambda}{2}\|w_c - w_{p,k}\|^2,
\label{eq:phase2_loss}
\end{equation}

where $\mathcal{L}_k^{c}(w_c; \mathcal{D}_k)$ denotes the total training loss of client $k$ in cluster $c$, and $\mathcal{L}_k^{\text{CE}}(w_c; \mathcal{D}_k)$ denotes the standard cross-entropy loss evaluated on client $k$'s local dataset. The parameter $w_c$ represents the cluster model parameters, and $w_{p,k}$ denotes the personal reference model of client $k$, initialized from the Phase~1 global model $w^{(T_0)}$ and kept fixed during Phase~2 and Phase~3. 

Although all reference models share the same initial weights, maintaining separate copies facilitates personalization which can be done later. The regularization term prevents the cluster model from drifting excessively from the global representation while allowing adaptation to cluster-level data. The hyperparameter $\lambda$ controls the strength of this regularization.

After local training, clients return their updated cluster models $w_{c,k}^{(t)}$ to the server. The cluster model is then updated using a data-size--weighted aggregation:

\begin{equation}
w_c^{(t+1)} =
\sum_{k \in \mathcal{C}_c}
\frac{|\mathcal{D}_k|}{|\mathcal{D}_c|}
w_{c,k}^{(t)},
\end{equation}

where $|\mathcal{D}_c|=\sum_{k\in\mathcal{C}_c}|\mathcal{D}_k|$ denotes the total number of samples within cluster $c$.
The cluster-level stabilization phase is performed for $T_1$ communication rounds, where $T_1 \geq 0$. When $T_1=0$, the cluster models are not further refined and are directly forwarded to the subsequent pruning phase.

\subsection{Phase 3: Cluster Training with Pruning}
Phase~3 continues optimizing the same regularized cluster objective defined in Eq.~\eqref{eq:phase2_loss} in Phase~2, while introducing pruning into the cluster models.

The cluster model $w_c$ is subjected to a binary pruning mask $M_c$ \cite{lin2022federated}. This step ensures that pruned weights remain zeroed out in the global cluster state:

\begin{equation}
w_c^{(t+1)} = \left( \sum_{k \in \mathcal{C}_c} \frac{|\mathcal{D}_k|}{|\mathcal{D}_c|} \, w_{c,k}^{(t)} \right) \odot M_c^{(t)}
\end{equation}
To determine which parameters should be retained or pruned under the binary mask while preserving cluster-level knowledge, we introduce a cluster-aware importance scoring.

\textbf{Importance Scoring:} Let $w_{c,j}$ denote the $j$-th scalar weight parameter of the cluster model $w_c$, and let $W_c$ denote the set of all scalar weight parameters of $w_c$. CA-AFP integrates cluster-level signals to guide pruning through importance score:

\begin{equation}
\text{Score}_{w_{c,j}} =
\alpha \cdot \text{Mag}_{w_{c,j}} +
\beta \cdot \text{Coh}_{w_{c,j}} +
\gamma \cdot \text{Con}_{w_{c,j}},
\label{eq:importance_score}
\end{equation}
where $\alpha$, $\beta$, and $\gamma$ are weighting coefficients satisfying $\alpha+\beta+\gamma=1$.

\textbf{Magnitude Score.} The normalized absolute magnitude of a weight:
\begin{equation}
\text{Mag}_{w_{c,j}} =
\frac{|w_{c,j}|}{\max_{u \in W_c} |u|}
\end{equation}

\textbf{Coherence Score.} Measures cross-client stability of a weight within a cluster:
\begin{equation}
\text{Coh}_{w_{c,j}} =
\frac{1}{1 + \mathrm{Var}_{k \in \mathcal{C}_c}(w_{c,j}^{(k)})}
\end{equation}
where $w_{c,j}^{(k)}$ denotes the value of weight $w_{c,j}$ in the locally trained cluster model of client $k$. Lower variance indicates higher structural agreement across clients.

\textbf{Consistency Score.} Measures gradient direction agreement among cluster clients:
\begin{equation}
\text{Con}_{w_{c,j}} =
\left|
\frac{1}{|\mathcal{C}_c|}
\sum_{k \in \mathcal{C}_c}
\mathrm{sign}\!\left(
\frac{\partial \mathcal{L}_k}{\partial w_{c,j}}
\right)
\right|
\end{equation}
where $\mathcal{L}_k$ denotes the local objective of client $k$.

Here, $\mathcal{C}_c$ represents the set of clients belonging to cluster $c$. The importance score favors parameters that are simultaneously large in magnitude, stable across clients, and consistently supported by gradient directions, thereby preserving both structural importance and learning dynamics during pruning.

CA-AFP dynamically changes the binary pruning mask as the rounds progress so that it can leverage the stablised weight updates in those rounds and improve its decision of pruning and retaining the models weights. For this, it follows the Prune and Heal Mechanism that we decribe below.\newline
\textbf{Prune-and-Heal Mechanism.} 
It is performed periodically using a pruning schedule during cluster training, following an iterative \emph{prune--heal} loop. At each pruning step, CA-AFP executes the following procedure:

\begin{enumerate}
    \item \textbf{Importance Re-evaluation:} Importance scores are recomputed using magnitude, coherence, and consistency signals to reflect the current training dynamics.
    
    \item \textbf{Scheduled Pruning:} Let $S^{(t)}$ denote the current sparsity level, defined as the fraction of pruned (zero-valued) weights in the cluster model at iteration $t$. Based on the current sparsity $S^{(t)}$ and the target sparsity $S_{\text{target}}$, a fraction of the lowest-ranked weights is removed. The pruning amount is automatically adjusted according to the remaining pruning steps and a user-defined prune rate, ensuring a smooth progression toward the target sparsity.
    
    \item \textbf{Gradient-Guided Regrowth:} To preserve the model’s ability to adapt during training, a subset of pruned connections with the highest gradient magnitudes is reactivated, enabling structural exploration and preventing irreversible information loss.
    
    \item \textbf{Healing Phase:} The masked sparse model is then trained for few rounds as per the schedule defined, allowing the remaining parameters to adapt and compensate before the next pruning step.
\end{enumerate}

\subsection{Phase 4: Client Fine-Tuning}
After the pruning process is completed, each client performs a local fine-tuning stage to recover potential accuracy loss introduced by pruning. During this stage, the final pruning mask $M_c$ is fixed and no further structural updates are allowed.

Let $\tilde{w}_k$ denote the locally fine-tuned sparse model of client $k$, initialized from its final cluster model after Phase~3. Each client optimizes $\tilde{w}_k$ under the sparsity constraint:

\begin{equation}
\min_{\tilde{w}_{k}} \; \mathcal{L}_k^{\mathrm{CE}}(\tilde{w}_{k}; \mathcal{D}_k)
\quad \text{s.t.} \quad
\tilde{w}_{k} \odot M_c = \tilde{w}_{k}.
\end{equation}

This constraint enforces that only the parameters corresponding to non-zero entries in the pruning mask $M_c$ are updated, while pruned parameters remain fixed at zero. Fine-tuning is performed for a small number of local epochs without any further communication or aggregation, allowing each client to adapt the sparse model to its local data distribution while preserving the final sparsity level.

%================================================================%

\section{Experimental Setup}
%================================================================%

\subsection{Dataset and Data Distribution}

We evaluate CA-AFP on two widely used human activity recognition (HAR) benchmarks, \textbf{WISDM} and \textbf{UCI-HAR}. These datasets are selected because they are standard benchmarks in federated and personalized HAR research \cite{aouedi2024federated}, exhibit strong inter-user heterogeneity, and differ in sensing conditions and data collection environments, enabling a robust assessment of CA-AFP.

\paragraph{WISDM Dataset:}
The WISDM (Wireless Sensor Data Mining) activity recognition dataset~\cite{kwapisz2011activity} consists of 1,098,207 raw smartphone accelerometer recordings collected from 36 users performing six activities: Walking, Jogging, Upstairs, Downstairs, Sitting, and Standing. Raw sensor streams are segmented using sliding windows of 200 timesteps with 50\% overlap, producing 10{,}591 samples of shape $(200,3)$. Each user is treated as an independent federated client, resulting in a natural user-based data partition that preserves real-world heterogeneity arising from differences in motion patterns, activity frequency, and device handling.

This natural partition that we summarise in Table \ref{tab:heterogeneity_comparison} inherently induces non-IID data distributions across clients, including feature skew (due to individual motion dynamics), label skew (uneven activity participation across users), and quantity skew (varying numbers of samples per user), without introducing artificial data manipulation. WISDM therefore serves as a realistic, noisy, and unconstrained federated learning benchmark.

\paragraph{UCI-HAR Dataset:}
The UCI-HAR dataset~\cite {garcia2020public} contains inertial sensor data from 30 subjects performing six activities under controlled laboratory conditions, collected using waist-mounted smartphones. The dataset comprises a total of 10,299 labeled samples, where each sample is represented as a multivariate time series of shape $(128,9)$. Similar to WISDM, we adopt a natural federated setting where each subject corresponds to one client. Compared to WISDM, UCI-HAR exhibits lower sensor noise and more consistent activity execution (Table \ref{tab:heterogeneity_comparison}).
\begin{table}[t]
\centering
\caption{Heterogeneity analysis of WISDM and UCI-HAR.}
\label{tab:heterogeneity_comparison}
\renewcommand{\arraystretch}{1.2}

\resizebox{\columnwidth}{!}{%
    \begin{tabular}{l l c c}
    \hline
    \textbf{Heterogeneity Type} & \textbf{Metric} & \textbf{WISDM} & \textbf{UCI HAR} \\
    \hline
    \textbf{Quantity Skew} & Coef. of Variation ($CV = \frac{\sigma}{\mu} \times 100 \% $) & 30.4\% & 10.2\% \\
     & Sample Count ($\mu \pm \sigma$) & $297 \pm 90$ & $343 \pm 35$ \\
    \hline
    \textbf{Label Skew} & Clients with Missing Classes & 17 / 36 & 0 / 30 \\
     & Missing Rate & 47.2\% & 0.0\% \\
    \hline
    \textbf{Feature Skew} & Mean Inter-Client CV & 6.5\% & 0.6\% \\
 %    & & \\
    \hline
    \end{tabular}%
}
\end{table}

Together, WISDM and UCI-HAR span diverse deployment conditions, including real-world versus laboratory environments, different window lengths and sensor dimensionalities, and varying degrees of inter-client heterogeneity. This combination enables a balanced evaluation of CA-AFP under both realistic and controlled federated learning settings.

\subsection{Baseline Methods}

We compare CA-AFP against four representative baselines that address federated heterogeneity through either client clustering or model pruning. FedCHAR and ClusterFL focus on clustering to mitigate statistical heterogeneity and improve client-level performance, while EfficientFL and FedSNIP employ pruning to enhance communication efficiency and device compatibility. In addition to reducing communication overhead, pruning introduces regularization that helps mitigate overfitting and client drift~\cite{huang2022fedmask}. We refer to methods without pruning as \textit{dense baselines} (FedCHAR, ClusterFL) and those with pruning as \textit{pruned baselines} (EfficientFL, FedSNIP).

\textbf{FedCHAR}~\cite{li2023hierarchical} is a clustering-based personalized FL method without pruning and serves as an accuracy-oriented upper bound.
\textbf{ClusterFL}~\cite{ouyang2023clusterfl} jointly optimizes client models and cluster assignments via alternating optimization to mitigate statistical heterogeneity.
\textbf{EfficientFL}~\cite{wu2023efficient} compresses client updates through sparsification and quantization to reduce communication cost while preserving convergence.
\textbf{FedSNIP}~\cite{bustincio2025reducing} applies one-shot SNIP-based pruning at clients to transmit only critical parameters.

Recent methods such as SAFL and FLCAP also combine clustering and pruning but operate under different architectural and algorithmic assumptions. SAFL relies on Batch Normalization–based network slimming, which is incompatible with our BN-free architecture. FLCAP derives clustering signals from pruning-mask dynamics rather than model update similarity. To ensure fair and controlled comparisons, we therefore restrict our evaluation to baselines that are directly compatible with our model and training.

\begin{table*}[t]
    \centering
    \caption{Performance comparison on the WISDM and UCI-HAR datasets with 3 and 25 fine-tuning (FT) epochs.}
    \label{tab:main_results_combined}
    \setlength{\tabcolsep}{4pt} 
    \begin{tabular}{l c c ccc ccc}
        \toprule
        & & & \multicolumn{3}{c}{\textbf{WISDM}} & \multicolumn{3}{c}{\textbf{UCI-HAR}} \\
        \cmidrule(lr){4-6} \cmidrule(lr){7-9}
        \textbf{Method} & \textbf{Sparsity} & \textbf{FT Epochs} & \textbf{Avg Acc ($\mu$)} & \textbf{Fairness ($\sigma$)} & \textbf{Comm} & \textbf{Avg Acc ($\mu$)} & \textbf{Fairness ($\sigma$)} & \textbf{Comm} \\
        & & & ($\uparrow$) & ($\downarrow$) & (MB) & ($\uparrow$) & ($\downarrow$) & (MB) \\
        \midrule
        \textit{Dense Baseline} & & & & & & & & \\
        FedCHAR~\cite{li2023hierarchical} & 0\% & 3 & 96.64\% & 8.39\% & 483 & 96.87\% & 6.70\% & 347 \\
        & 0\% & 25 & 97.16\% & 8.32\% & 483 & 97.48\% & 6.19\% & 347 \\
        ClusterFL~\cite{ouyang2023clusterfl} & 0\% & 3 & 96.18\% & 7.02\% & 474 & 95.80\% & 6.87\% & 340 \\
        & 0\% & 25 & 96.33\% & 6.50\% & 474 & 95.80\% & 6.87\% & 340 \\
        \midrule
        \textit{Pruned Baselines} & & & & & & & & \\
        EfficientFL~\cite{wu2023efficient} & $\approx$ 70\% & 3 & 92.30\% & 10.29\% & 142 & 95.02\% & 9.46\% & 102 \\
        & $\approx$ 70\% & 25 & 95.68\% & 9.28\% & 142 & 97.49\% & 5.85\% & 102 \\
        FedSnip~\cite{bustincio2025reducing} & $\approx$ 70\% & 3 & 94.29\% & 10.16\% & 142 & 95.56\% & 9.34\% & 102 \\
        & $\approx$ 70\% & 25 & 95.24\% & 8.96\% & 142 & 95.82\% & 9.34\% & 102 \\
        \midrule
        \textit{Ours} & & & & & & & & \\
        \textbf{CA-AFP (0,0,50)} & $\approx$ 70\% & 3 & 95.56\% & 8.97\% & 176 & 95.99\% & 6.91\% & 122 \\
        & $\approx$ 70\% & 25 & 95.62\% & 8.84\% & 176 & 97.13\% & 6.55\% & 122 \\
        \textbf{CA-AFP (15,15,20)} & $\approx$ 70\% & 3 & 95.28\% & 9.50\% & 375 & 95.91\% & 6.49\% & 266 \\
        & $\approx$ 70\% & 25 & 96.73\% & 8.35\% & 375 & 97.40\% & 6.40\% & 266 \\
        \bottomrule
       % \vspace{-2em}
    \end{tabular}
\end{table*}

\subsection{Experimental settings}
\label{sec:exp_settings}
We describe our experimental configurations for reproducibility. Additional details are provided in Appendix (Tables~\ref{tab:uci_hyperparams}--\ref{tab:noniid_setup}).

\paragraph{Implementation and Hardware.}
All experiments were conducted on a Linux workstation equipped with an Intel Core i9-14900K processor. CA-AFP and all baselines were implemented in TensorFlow~2.14 and initialized using a fixed random seed for both weight initialization and client data partitioning. Unless otherwise stated, we use Adam with a learning rate of $10^{-3}$. The target sparsity is fixed at 70\%, and all methods are trained for 50 global rounds. Reported results in  Table~\ref{tab:main_results_combined} are averaged over five random seeds.

\paragraph{Phase Sensitivity Analysis.}
To isolate the effects of different training phases, we conduct sensitivity analyses on the WISDM dataset. We vary the durations of the initial training phase ($P_1$) and post-clustering phase ($P_2$) from 0 to 20 rounds, and the starting sparsity ($S_{\text{start}}$) from 0.0 to 0.9 (Table~\ref{tab:ablation_phases}). The pruning phase is fixed at $P_3=40$ rounds, and fine-tuning is set to $P_4=3$ epochs. Results are averaged over three random seeds.

\paragraph{Importance Score Ablation.}
We analyze the importance scoring function on the UCI-HAR  by evaluating seven combinations of magnitude ($\alpha$), coherence ($\beta$), and consistency ($\gamma$) weights. The pruning schedule is varied from 30\% to 70\% over 15 rounds to examine the contribution of each component. The configurations are reported in Table~\ref{tab:score_ablation_setup}. These experiments are conducted using a single random seed.

\paragraph{Non-IID Robustness Evaluation.}
We evaluate all methods on WISDM under three levels of statistical heterogeneity (Table~\ref{tab:noniid_setup}). Each client is restricted to at most $k$ activity classes: \textit{Extreme} ($k=1$), \textit{Moderate} ($k=2$), and \textit{Mild} ($k=3$). Allowed classes are randomly sampled from each client’s ground-truth labels, and data from other classes are discarded. All experiments run for 50 global rounds with target sparsity $S=0.7$, and results are reported as mean $\pm$ standard deviation over five seeds.

\paragraph{Fine-Tuning Setup.}
Fine-tuning configurations are summarized in Table~\ref{tab:finetune_config}.

\paragraph{Model Architecture.}
All methods use the same lightweight 1D CNN for human activity recognition. The network consists of two convolutional blocks, each containing a 1D convolution layer with 64 filters and kernel size 5, followed by ReLU activation, max pooling, and dropout (rate 0.3). The resulting features are flattened and passed to a fully connected layer with 32 units and ReLU activation, followed by dropout (rate 0.2) and a Softmax output layer.

\subsection{Accuracy metrics}

To compare with the existing baselines, and to analyse ablations, we use three metrics, namely Accuracy, Fairness (Standard Deviation) and Communication Cost (in MB).
The average Accuracy is the mean test accuracy ($\mu$) across all clients. We use the term `fairness' to refer to the standard deviation ($\sigma$) of the mean accuracy across all clients. A lower value of fairness means the test accuracy differs less between different clients which indicates that the models had balanced performance across all the clients whereas a higher value for fairness means that the test accuracy varied more between different clients which indicates that the model is not reliable across all clients as its performance varies more. The communication cost measures the total volume of data transmitted between the server and clients throughout the entire training process, expressed in Megabytes (MB). It accounts for both client-to-server and server-to-client transfers, reflecting efficiency gains from sparsity. The total cost $C_{total}$ over $R$ rounds is estimated as:
\begin{equation}
    C_{total} = \sum_{r=1}^{R} \sum_{k \in \mathcal{S}_r} 2 \times |W| \times (1 - S_k^{(r)}) \times \frac{4}{1024^2}
\end{equation}
where $\mathcal{S}_r$ is the set of selected clients in round $r$, $|W|$ is the total number of parameters in the dense model (we consider it 32-bit floats, i.e., 4 bytes), and $S_k^{(r)}$ is the sparsity of client $k$'s model in round $r$. 

\begin{table}[]
\centering
\caption{Best ``importance score'' configurations.}
\label{tab:alpha_beta_gamma_ablation_final}
% Resize the table to fit the column width
\resizebox{\columnwidth}{!}{%
    \begin{tabular}{l l c c c c c}
    \toprule
    \textbf{Scenario} & \textbf{Dataset} & $\alpha$ & $\beta$ & $\gamma$ & \textbf{Avg Acc (\%)} & \textbf{Fairness (\%)} \\
    \midrule
    Standard & UCI HAR & 0.25 & 0.25 & 0.50 & 91.19 & 9.81 \\
             & WISDM   & 0.25 & 0.25 & 0.50 & 91.57 & 10.51 \\
    \midrule
    Drift & UCI HAR & 0.25 & 0.25 & 0.50 & 66.54 & 32.70 \\
          & WISDM   & 0.00 & 0.00 & 1.00 & 57.63 & 33.12 \\
    \midrule
    Noisy Clients & UCI HAR & 0.50 & 0.25 & 0.25 & 74.42 & 19.03 \\
                  & WISDM   & 0.25 & 0.25 & 0.50 & 74.36 & 20.72 \\
    \bottomrule
    \end{tabular}%
}
\end{table}

\begin{figure*}[!t]
    \centering
    \includegraphics[width=\textwidth]{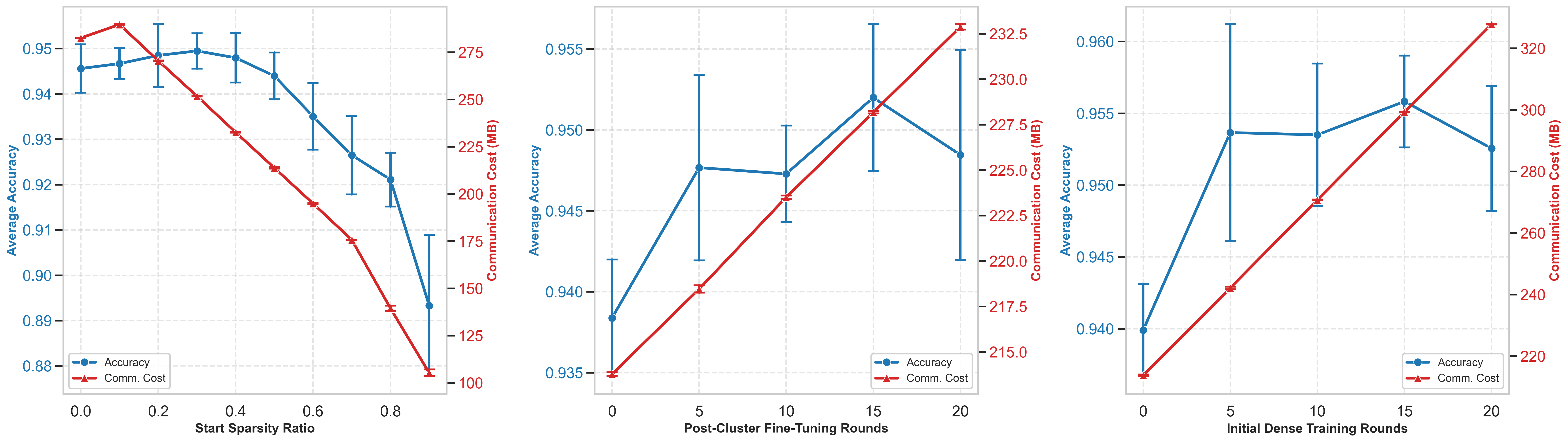}
    \caption{Effect of ``Start Sparsity'' and ``Rounds'' on accuracy in WISDM.} 
    \label{fig:method_workflow}
    \vspace{-1em}
\end{figure*}

%================================================================%
\section{Results}
%================================================================%

Table~\ref{tab:main_results_combined} gives the performance comparison of different baseline methods and CA-AFP across both datasets. 
From this table, we observe that there is tradeoff between the accuracy and the communication cost. Dense methods such as FedCHAR and ClusterFL achieve strong accuracy but incur the highest communication overhead due to full-model communication. In contrast, pruning-based approaches significantly reduce communication cost by transmitting sparse updates.
The performance is generally more equitable across the clients as the accuracy increases but in some cases it does not show this relation.
CA-AFP achieves a better balance between accuracy, efficiency, and fairness. It consistently performs better or at par with the pruning  baselines and with the (15, 15, 20) rounds configuration, it also performs competitively with the dense baselines while incurring lesser communication cost.  The CAAFP with (0, 0, 50) rounds configuration has slightly higher communication cost as compared to the other pruning baselines with because it also regrows weights at a defined rate while pruning. However, the extra cost remains constant given a dataset and model size.

Together, these results convey that CA-AFP simultaneously aims for objectives: high accuracy, low communication cost, and improved inter-client fairness. It also gives the flexibility utilise the communication budget by adjusting the rounds distribution among different phases to acheive a given target sparisty level for deployment of the models.

\subsection{\textbf{Ablation and Sensitivity Results}}
\begin{comment}
We experiment on four different aspects of the CA-AFP method.
From Eq\ref{eq:importance_score} we used coefficients $(\alpha,\beta,\gamma)$ to weight the threee different terms in the importance score equation, but we need to identify whether all three terms are useful or not. For that, we perform an experiment in three different FL scenarios with different combinations of these weights. 
We isolate and analyse the effect of the three hyperparameters in CA-AFP, namely the  number of initial training rounds before pruning and clustering, the number of training rounds after clustering but before pruning, and the sparisty level at which the pruning begins.
We also conduct the experiment to understand the effect of Fine-tuning rounds on CA-AFP, Fedsnip,EfficientFL,FedCHAR and a variant CA-AFP(Global-FT) where we make only one cluster(i.e no clustering).
We also evaluate our method and the baselines under different levels of non iid. The specfic connfiguration are mentioned in the Table \ref{tab:noniid_setup}
\end{comment}

\subsubsection{Weights of Importance Score}
We conduct an ablation study to analyze the effect of the importance score weights $(\alpha,\beta,\gamma)$ under four heterogeneous data scenarios: standard Non-IID,  noisy clients, and drift.

\noindent \textbf{Scenario 1: Standard (Natural Heterogeneity).} This scenario serves as the baseline, utilizing the natural partitioning of the UCI HAR dataset. Data is distributed such that each federated learning client corresponds strictly to a single human subject from the original study, preserving the inherent user-specific variations without synthetic modification.

\noindent \textbf{Scenario 2: Noisy Clients (Partial Label Corruption).} This scenario simulates unreliable data annotation. A random subset of $40\%$ of the clients is selected to act as noisy participants. For these clients, $30\%$ of their labels are replaced with random integers drawn from a uniform discrete distribution $U(0, 5)$, while the feature vectors $X$ remain unchanged.

\noindent \textbf{Scenario 3: Drift (Sensor Malfunction).} This scenario simulates sensor failure or broken devices. In this setting, $40\%$ of the clients are selected to contribute pure noise. For these participants, $100\%$ of the labels are replaced with random values, effectively destroying the mapping between features and labels ($X \rightarrow Y_{random}$). This stress-tests the importance scoring mechanism's ability to identify and suppress updates from malfunctioning nodes.

The best-performing configuration for each scenario is summarized in Table~\ref{tab:alpha_beta_gamma_ablation_final}.
Overall, these results demonstrate that no single component of importance score dominates across all environments. Instead, the mixed formulation enables CA-AFP to flexibly adapt pruning behavior to different types of data and its modifications. This validates our design choice of using a weighted hybrid importance score rather than relying on any single pruning criterion.

The complete ablation results for all weight combinations and scenarios are reported in Table ~\ref{tab:ablation_combined} in Appendix. The combinations having highest accuracy to fairness ratio which we term as 'score' in Table \ref{tab:ablation_combined} is selected as higher accuracy and lower value of fairness is better.

\subsubsection{Effect of Rounds and Start Sparsity}
In this study, we experiment to isolate and understand the effect of number of rounds for initial training (of phase 1), number of rounds after clustering but before pruning (phase 2) and the starting sparsity level of the pruning and heal (phase 3) with respect to the communication cost and average accuracy across the clients. From the results (see Figure~\ref{fig:method_workflow}), it is observed that increase in accuracy always comes with an increase in the communication cost. Lowering the starting sparsity as well as increasing the number of initial training rounds in phase 1 or rounds in phase 2 is effective in increasing the accuracy but with increase in the communication cost. We also observe that after increasing the starting sparsity beyond a certain value (in this case, 0.5), the accuracy drops steeply while the communication cost decreases almost at a constant rate. So it is useful to find this number for start sparsity to get better results with limited cost.

\begin{figure}[t]
    \centering
    \includegraphics[width=0.9\linewidth]{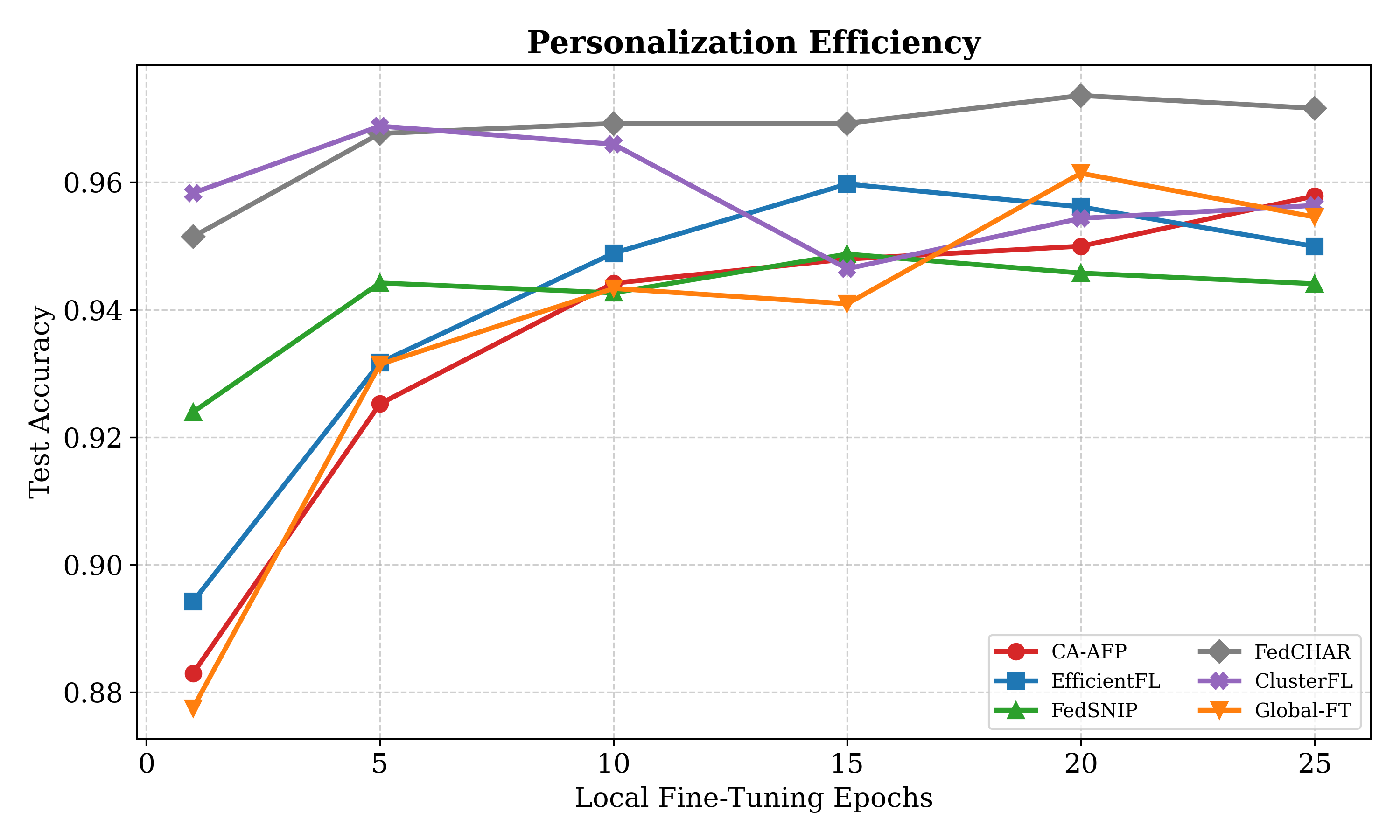}
    \caption{Impact of Local Fine-Tuning on accuracy in WISDM.}
    \label{fig:finetune_efficacy}
    \vspace{-2em}
\end{figure}

\subsubsection{Effect of Local Fine-Tuning}
\label{sec:finetuning_analysis}

In this experiment, we analyze the impact of local fine-tuning epochs (Phase 4) on final ``Accuracy''. Since fine-tuning is performed locally on the sparse models, it incurs zero additional communication overhead.

Increasing the number of local fine-tuning epochs from 0 to 25 improves the accuracy of CA-AFP from approximately 88\% to 96\%, as shown in Figure~\ref{fig:finetune_efficacy}, while maintaining a constant communication cost of approximately 176 MB. This demonstrates that the parameters retained by CA-AFP remain highly adaptable, enabling clients to recover performance through local fine-tuning without additional communication overhead.

The \textit{Global-FT} baseline in Figure~\ref{fig:finetune_efficacy}, which corresponds to CA-AFP with a single cluster, consistently achieves lower accuracy than clustered CA-AFP in the early fine-tuning stages. This indicates that clustering facilitates earlier learning of personalized representations. Moreover, using 10–25 local fine-tuning epochs effectively mitigates any performance gap introduced by cluster-based data partitioning, resulting in stable and robust convergence. Consistent trends are also observed in Table~\ref{tab:main_results_combined}, where fine-tuning uniformly improves CA-AFP performance across experimental settings.

\begin{figure}[t]
    \label{fig:non_iid}
    \centering
    \includegraphics[width=1.0\linewidth]{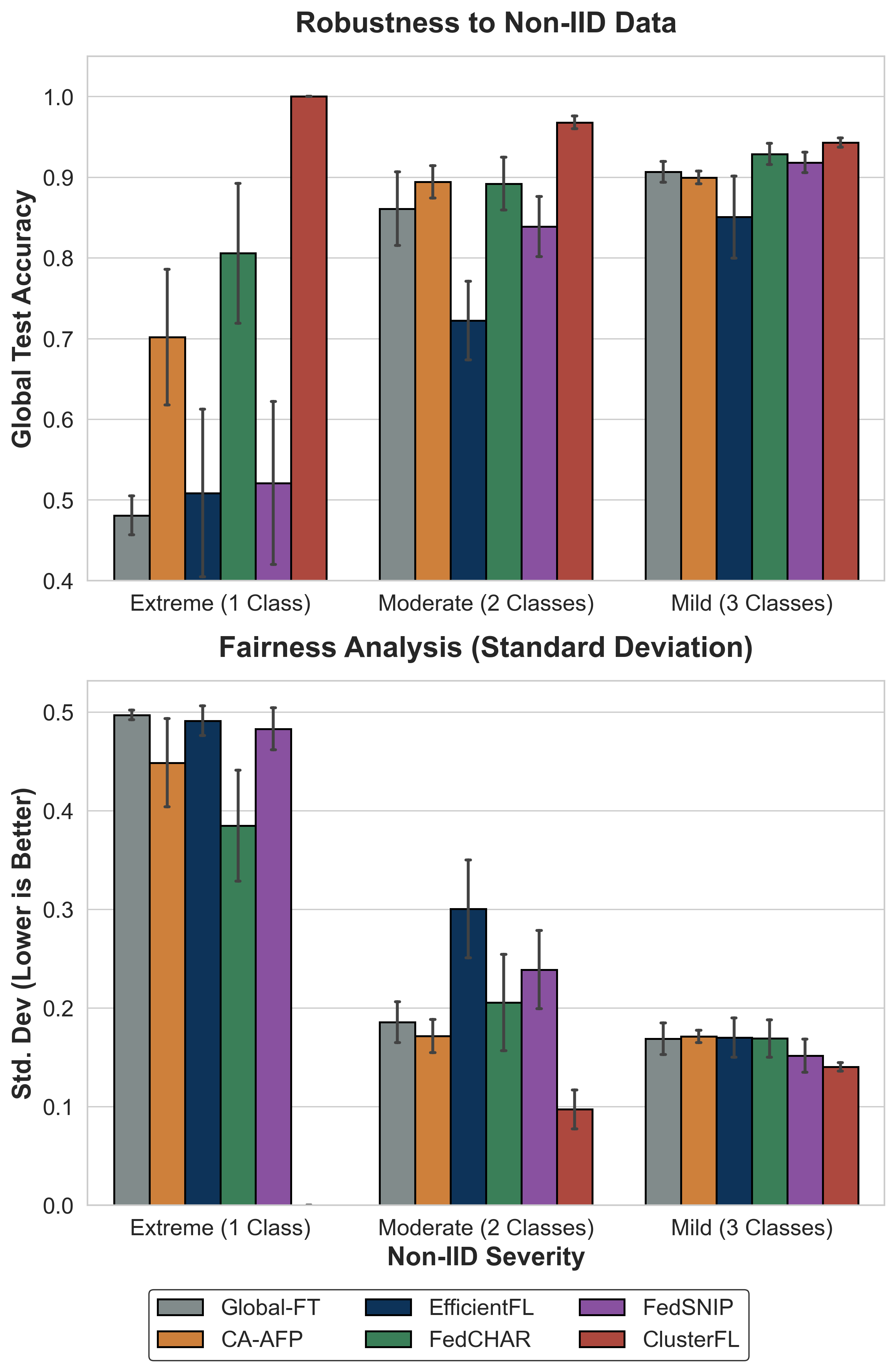}
    \caption{Performance across three Non-IID levels.}
    \label{fig:scenario_comparison}
    \vspace{-2em}
\end{figure}

\subsubsection{Robustness across Non-IID levels}
\label{sec:scenario_analysis}

We evaluate the robustness of CA-AFP and baseline methods under three levels of statistical heterogeneity: \textit{Mild} (3 classes per client), \textit{Moderate} (2 classes per client), and \textit{Extreme} (1 class per client).

As shown in Figure~\ref{fig:scenario_comparison}, under moderate and extreme non-IID settings, pruning-based baselines exhibit substantially lower accuracy than clustered dense methods and CA-AFP. In particular, under the extreme 1-class-per-client setting, pruning baselines (FedSNIP and EfficientFL) degrade to approximately 50\% accuracy, largely due to severe gradient conflicts across heterogeneous clients.

Since no fine-tuning is performed in this experiment, these pruning methods are unable to adapt to client-specific distributions, as they rely on a single global model without clustering. In contrast, CA-AFP mitigates this issue by isolating conflicting objectives into separate clusters, maintaining an accuracy of approximately 80% while preserving communication efficiency.

Similar trends are observed under the moderate (2-class) non-IID setting, where pruning baselines continue to underperform relative to clustering-based approaches and CA-AFP. Furthermore, the CA-AFP variant without clustering (\textit{Global-FT}) performs poorly under extreme heterogeneity but recovers accuracy under moderate heterogeneity, highlighting the critical role of clustering in handling severe client-level data diversity.

%================================================================%
\section{Limitations and Future Work}
%================================================================%

\textbf{Clustering Strategies.}
In this work, we adopt a fixed clustering strategy to balance effectiveness and computational efficiency. Although prior studies have explored alternative clustering mechanisms~\cite{li2023hierarchical,islam2024fedclust}, a systematic analysis of how different clustering metrics, schedules, and integration points interact with adaptive pruning remains an interesting direction for future work. 

\textbf{Hyperparameter Sensitivity.}
CA-AFP relies on a small number of hyperparameters, including the importance score weights and the global sparsity target ($S_{\text{target}}=70\%$). Our ablation studies demonstrate stable performance across a wide range of settings. Nevertheless, developing adaptive mechanisms that relate these parameters to underlying data distributions could further reduce tuning effort and improve usability in large-scale deployments.

%================================================================%
%================================================================%
\section{Conclusion}
%================================================================%

In this work, we presented CA-AFP, a unified framework that integrates clustering-based personalization with adaptive model pruning for federated learning under heterogeneous conditions. Through two key innovations—a cluster-aware importance scoring mechanism that combines weight magnitude, intra-cluster coherence, and gradient consistency, and an iterative self-healing pruning schedule—CA-AFP achieves a favorable balance between predictive accuracy, inter-client fairness, and communication efficiency.

Extensive experimental results demonstrate that CA-AFP remains robust under extreme non-IID settings and enables flexible adaptation to varying communication budgets through controllable pruning and fine-tuning schedules. By prioritizing parameters based on cluster-level consensus rather than isolated client statistics, CA-AFP makes federated learning more practical for deployment on resource-constrained edge devices. Overall, our work highlights the potential of jointly optimizing personalization and resource efficiency for scalable and reliable federated systems.

%================================================================%

\bibliography{sample-base}
\bibliographystyle{acm}
\clearpage

\section{Appendix}

\subsection{Algorithm}
In this section we have explained the CA-AFP framework through the psuedo code as mentioned in the Algorithm \ref{alg:caafp_full}
\begin{algorithm}
\caption{CA-AFP: Cluster-Aware Adaptive Federated Pruning}
\label{alg:caafp_full}
\begin{algorithmic}[1]\footnotesize

\Require Clients $\mathcal{K}$, clusters $C$, warm-up rounds $T_1\!\ge\!0$, 
stabilization rounds $T_2\!\ge\!0$, pruning rounds $T_3$, pruning frequency $f$, 
target sparsity $S_{\text{target}}$, prune rate $\rho$

\Ensure Personalized models $\{v_k\}$

\State Initialize global model $w^{(0)}$

% ---------------- PHASE 1 ----------------
\Statex
\State \textbf{Phase 1: Warm-up Training}

\If{$T_1>0$}
\For{$t=1$ to $T_1$}
    \State Server broadcasts $w^{(t)}$
    \ForAll{clients $k$}
        \State Client trains locally and returns $w_k^{(t)}$
    \EndFor
    \State $w^{(t+1)} \gets \sum_k \frac{|\mathcal{D}_k|}{|\mathcal{D}|} w_k^{(t)}$
\EndFor
\EndIf

% ---------------- CLUSTERING ----------------
\Statex
\State \textbf{Client Clustering (Always Executed)}

\State Let $w^{\text{ref}} \gets w^{(T_1)}$ \Comment{$w^{(0)}$ if $T_1=0$}

\ForAll{clients $k$}
    \State Client trains one local epoch from $w^{\text{ref}}$
    \State Compute update $\Delta w_k$
\EndFor

\State Server clusters $\{\Delta w_k\}$ into $C$ clusters

\ForAll{clusters $c$}
    \State $w_c^{(0)} \gets w^{\text{ref}}$
    \State $M_c^{(0)} \gets \mathbf{1}$
\EndFor

% ---------------- PHASE 2 ----------------
\Statex
\State \textbf{Phase 2: Dense Cluster Stabilization}

\If{$T_2>0$}
\For{$t=1$ to $T_2$}
\ForAll{clusters $c$}
    \State Server broadcasts $w_c^{(t)}$
    \ForAll{clients $k\in\mathcal{C}_c$}
        \State Client minimizes $\mathcal{L}_k^c(w_c)$ and returns $w_{c,k}^{(t)}$
    \EndFor
    \State $w_c^{(t+1)} \gets 
    \sum_{k\in\mathcal{C}_c}\frac{|\mathcal{D}_k|}{|\mathcal{D}_c|} w_{c,k}^{(t)}$
\EndFor
\EndFor
\EndIf

% ---------------- PHASE 3 ----------------
\Statex
\State \textbf{Phase 3: Prune--Heal Training}

\For{$t=1$ to $T_3$}
\ForAll{clusters $c$}

\If{$t \bmod f = 0$}
    \State Clients compute gradients using $w_c^{(t)} \odot M_c^{(t)}$
    \State Compute importance scores 
    $\text{Score}_w=\alpha\text{Mag}_w+\beta\text{Coh}_w+\gamma\text{Con}_w$
    \State Compute sparsity $S^{(t)}$
    \State $R \gets \lfloor (T_3-t)/f \rfloor$
    \State $\Delta S \gets (S_{\text{target}}-S^{(t)})/R$
    \State $N_{\text{churn}} \gets \rho \cdot N_{\text{active}}$
    \State $N_{\text{deficit}} \gets \Delta S \cdot N_{\text{total}}$
    \State $N_{\text{prune}} \gets N_{\text{churn}}+N_{\text{deficit}}$
    \State $N_{\text{grow}} \gets N_{\text{churn}}$
    \State Prune $N_{\text{prune}}$ lowest-score active weights
    \State Regrow $N_{\text{grow}}$ highest-gradient inactive weights
    \State Update mask $M_c^{(t)}$
\EndIf

\State Server broadcasts $(w_c^{(t)},M_c^{(t)})$

\ForAll{clients $k\in\mathcal{C}_c$}
    \State Minimize $\mathcal{L}_k^c(w_c)$ under mask $M_c^{(t)}$
    \State Return $w_{c,k}^{(t)}$
\EndFor

\State $w_c^{(t+1)} \gets 
\left(\sum_{k\in\mathcal{C}_c}\frac{|\mathcal{D}_k|}{|\mathcal{D}_c|}
w_{c,k}^{(t)}\right)\odot M_c^{(t)}$

\EndFor
\EndFor

% ---------------- PHASE 4 ----------------
\Statex
\State \textbf{Phase 4: Personalized Fine-Tuning (Optional)}

\ForAll{clusters $c$}
    \State $w_c^* \gets w_c^{(T_3)} \odot M_c^{(T_3)}$
    \ForAll{clients $k\in\mathcal{C}_c$}
        \State $v_k \gets w_c^*$
        \State Client minimizes 
        $\mathcal{L}_k^{CE}(v_k;\mathcal{D}_k)$
        \Statex \hspace{1.2cm}\textbf{subject to } $v_k \odot M_c^{(T_3)} = v_k$
    \EndFor
\EndFor

\State \Return $\{v_k\}$
\end{algorithmic}
\end{algorithm}

\subsection{Complete Ablation Results for Score coefficients}

Table~\ref{tab:ablation_combined} reports the complete ablation results of the importance score under all evaluated weight configurations and dataset scenarios. 

\begin{table*}[t]
\centering
\caption{Combined ablation study of importance score weights $(\alpha, \beta, \gamma)$ on UCI HAR and WISDM datasets. \textbf{Note:} "Extreme Non-IID" is excluded as it was not evaluated on WISDM.}
\label{tab:ablation_combined}
\setlength{\tabcolsep}{4pt} % Adjust column spacing to fit width
\begin{tabular}{l ccc ccc ccc}
\toprule
 & \multicolumn{3}{c}{\textbf{Weights}} & \multicolumn{3}{c}{\textbf{UCI HAR}} & \multicolumn{3}{c}{\textbf{WISDM}} \\
\cmidrule(lr){2-4} \cmidrule(lr){5-7} \cmidrule(lr){8-10}
\textbf{Scenario} & $\alpha$ & $\beta$ & $\gamma$ & \textbf{Acc} & \textbf{Fairness} & \textbf{Score} & \textbf{Acc} & \textbf{Fairness} & \textbf{Score} \\
\midrule
Drift & 1.00 & 0.00 & 0.00 & 0.6860 & 0.3526 & 1.9453 & 0.6129 & 0.3820 & 1.6047 \\
Drift & 0.00 & 1.00 & 0.00 & 0.6339 & 0.3148 & 2.0135 & 0.5915 & 0.3456 & 1.7116 \\
Drift & 0.00 & 0.00 & 1.00 & 0.6475 & 0.3264 & 1.9840 & 0.5763 & 0.3312 & 1.7399 \\
Drift & 0.50 & 0.25 & 0.25 & 0.6939 & 0.3452 & 2.0102 & 0.5774 & 0.3396 & 1.7001 \\
Drift & 0.25 & 0.50 & 0.25 & 0.6853 & 0.3546 & 1.9323 & 0.5974 & 0.3671 & 1.6273 \\
Drift & 0.25 & 0.25 & 0.50 & 0.6654 & 0.3270 & 2.0350 & 0.6169 & 0.3671 & 1.6807 \\
Drift & 0.33 & 0.33 & 0.34 & 0.6967 & 0.3506 & 1.9870 & 0.5797 & 0.3420 & 1.6950 \\
\midrule
Standard & 1.00 & 0.00 & 0.00 & 0.8964 & 0.1550 & 5.7816 & 0.9143 & 0.1037 & 8.8198 \\
Standard & 0.00 & 1.00 & 0.00 & 0.8548 & 0.1416 & 6.0361 & 0.9040 & 0.1019 & 8.8708 \\
Standard & 0.00 & 0.00 & 1.00 & 0.8139 & 0.1549 & 5.2535 & 0.8906 & 0.1211 & 7.3559 \\
Standard & 0.50 & 0.25 & 0.25 & 0.8796 & 0.1691 & 5.2014 & 0.9031 & 0.1095 & 8.2486 \\
Standard & 0.25 & 0.50 & 0.25 & 0.8971 & 0.1396 & 6.4262 & 0.8939 & 0.1152 & 7.7587 \\
Standard & 0.25 & 0.25 & 0.50 & 0.9119 & 0.0981 & 9.2963 & 0.9157 & 0.1051 & 8.7110 \\
Standard & 0.33 & 0.33 & 0.34 & 0.8769 & 0.1571 & 5.5809 & 0.9103 & 0.1145 & 7.9534 \\
\midrule
Noisy Clients & 1.00 & 0.00 & 0.00 & 0.7565 & 0.2111 & 3.5833 & 0.7456 & 0.2204 & 3.3832 \\
Noisy Clients & 0.00 & 1.00 & 0.00 & 0.6890 & 0.2006 & 3.4351 & 0.7449 & 0.2341 & 3.1822 \\
Noisy Clients & 0.00 & 0.00 & 1.00 & 0.6449 & 0.1800 & 3.5821 & 0.7322 & 0.2082 & 3.5170 \\
Noisy Clients & 0.50 & 0.25 & 0.25 & 0.7442 & 0.1903 & 3.9101 & 0.7490 & 0.2333 & 3.2100 \\
Noisy Clients & 0.25 & 0.50 & 0.25 & 0.7458 & 0.1949 & 3.8267 & 0.7335 & 0.2164 & 3.3894 \\
Noisy Clients & 0.25 & 0.25 & 0.50 & 0.6860 & 0.1963 & 3.4957 & 0.7436 & 0.2072 & 3.5893 \\
Noisy Clients & 0.33 & 0.33 & 0.34 & 0.7130 & 0.1843 & 3.8680 & 0.7560 & 0.2189 & 3.4542 \\
\bottomrule
\end{tabular}
\end{table*}
\subsection{Experimental Setup}

In this section, we present the configurations used in our experiments, detailed in Table \ref{tab:uci_hyperparams} - \ref{tab:noniid_setup}.

\begin{table}[H] % [H] forces the table to appear exactly HERE
\centering
\caption{Hyperparameter Configuration for UCI HAR Experiments}
\label{tab:uci_hyperparams}
\resizebox{\columnwidth}{!}{% % Ensures it fits within the column width
\renewcommand{\arraystretch}{1.1} % Slightly reduced for compactness
\begin{tabular}{l l c}
\toprule
\textbf{Method} & \textbf{Parameter} & \textbf{Value} \\
\midrule
\textbf{Common Settings} & Rounds phase wise ($P_1, P_2, P_3$) & 0, 0, 50 \\
 & Local Epochs ($E$) & 3 \\
 & Batch Size ($B$) & 32 \\
 & Optimizer & Adam ($lr=1e^{-3}$) \\
 & Fine-Tuning Rounds ($P_4$) & 3 \\
\midrule
\textbf{CA-AFP (Ours)} & Target Sparsity & Adaptive (Start 70\%) \\
 & Scoring Weights ($\alpha, \beta, \gamma$) & 0.5, 0.25, 0.25 \\
 & Mask Update Rate (Churn) & 5\% \\
 & Number of Clusters ($K$) & 3 \\
\midrule
\textbf{FedSNIP} & Target Sparsity & 70\% \\
\midrule
\textbf{EfficientFL} & Target Sparsity & 70\% \\
\midrule
\textbf{FedCHAR} & Clusters ($K$) & 3 \\
 & Warmup Rounds & 10 \\
\midrule
\textbf{ClusterFL} & Clusters ($K$) & 3 \\
 & Regularization ($\rho$) & $5e^{-3}$ \\
 & Bi-level Params ($\alpha, \beta$) & $1e^{-3}, 5e^{-4}$ \\
\bottomrule
\end{tabular}%
}
\end{table}

\begin{table}[htbp]
\centering
\caption{Configuration for WISDM Experiments}
\label{tab:wisdm_hyperparams}
\resizebox{\columnwidth}{!}{%
\renewcommand{\arraystretch}{1.2}
\begin{tabular}{l l c}
\toprule
\textbf{Method} & \textbf{Parameter} & \textbf{Value} \\
\midrule
\textbf{Common Settings} & Rounds phase wise ($P_1, P_2, P_3$) & 0, 0, 50 \\
 & Local Epochs ($E$) & 3 \\
 & Batch Size ($B$) & 32 \\
  & Optimizer & Adam ($lr=1e^{-3}$) \\
& Fine-Tuning Rounds ($P_4$) & \textbf{25} \\
\midrule
\textbf{CA-AFP (Ours)} & Target Sparsity & Adaptive (Start 70\%) \\
 & Scoring Weights ($\alpha, \beta, \gamma$) & 0.25, 0.25, 0.50 \\
 & Mask Update Rate (Churn) & 5\% \\
 & Number of Clusters ($K$) & 3 \\
\midrule
\textbf{FedSNIP} & Target Sparsity & 70\% \\
\midrule
\textbf{EfficientFL} & Target Sparsity & 70\% \\
\midrule
\textbf{FedCHAR} & Clusters ($K$) & 3 \\
 & Warmup Rounds & 10 \\
\midrule
\textbf{ClusterFL} & Clusters ($K$) & 3 \\
 & Regularization ($\rho$) & $5e^{-3}$ \\
 & Bi-level Params ($\alpha, \beta$) & $1e^{-3}, 5e^{-4}$ \\
\bottomrule
\end{tabular}%
}
\end{table}
\begin{table}[h]
\centering
\caption{Phase Configuration for CA-AFP Ablation Studies (WISDM)}
\label{tab:ablation_phases}
\resizebox{\columnwidth}{!}{%
\renewcommand{\arraystretch}{1.2}
\begin{tabular}{l | c c c c | c}
\toprule
\textbf{Ablation Study} & \textbf{$P_1$} & \textbf{$P_2$} & \textbf{$P_3$} & \textbf{$P_4$} & \textbf{Start} \\
 & \textbf{(Warmup)} & \textbf{(Adapt)} & \textbf{(Prune)} & \textbf{(Fine-Tune)} & \textbf{Sparsity} \\
\midrule
\textbf{Study A: Start Sparsity} & 0 & 0 & 40 & \textbf{3} & \textit{Varied} \\
\textit{(Varying $S_{start}$)} & & & & & ($0.0 - 0.9$) \\
\midrule
\textbf{Study B: Post-Cluster} & 0 & \textbf{\textit{Varied}} & 40 & \textbf{3} & 0.5 \\
\textit{(Varying $P_2$)} & & $\{0, 5, \dots, 20\}$ & & & (50\%) \\
\midrule
\textbf{Study C: Initial Warmup} & \textbf{\textit{Varied}} & 0 & 40 & \textbf{3} & 0.5 \\
\textit{(Varying $P_1$)} & $\{0, 5, \dots, 20\}$ & & & & (50\%) \\
\bottomrule
\end{tabular}%
}
\end{table}
\begin{table}[h]
\centering
\caption{Configuration for Importance Score Ablation Study}
\label{tab:score_ablation_setup}
\resizebox{\columnwidth}{!}{%
\renewcommand{\arraystretch}{1.2}
\begin{tabular}{l l l}
\toprule
\textbf{Component} & \textbf{Parameter} & \textbf{Value / Description} \\
\midrule
\textbf{Scoring Weights} & \textbf{Grid Search} & $(1.0, 0.0, 0.0)$ -- \textit{Pure Magnitude} \\
\textbf{($\alpha, \beta, \gamma$)} & (7 Combinations) & $(0.0, 1.0, 0.0)$ -- \textit{Pure Coherence} \\
& & $(0.0, 0.0, 1.0)$ -- \textit{Pure Consistency} \\
& & $(0.50, 0.25, 0.25)$ -- \textit{Magnitude Dominant} \\
& & $(0.25, 0.50, 0.25)$ -- \textit{Coherence Dominant} \\
& & $(0.25, 0.25, 0.50)$ -- \textit{Consistency Dominant} \\
& & $(0.33, 0.33, 0.34)$ -- \textit{Balanced} \\
\midrule
\textbf{Data Scenarios} & \textbf{Standard} & Natural User Split of UCI HAR and WISDM \\
& \textbf{Drift} &  40\% clients with 100\% Random labels \\ && (Sensor Malfunction)  \\
& \textbf{Noisy Clients} & 40\% clients with 30\% random labels \\
\midrule
\textbf{Sparsity Schedule} & Start Sparsity ($S_{start}$) & \textbf{30\%} (0.3) \\
& Target Sparsity ($S_{target}$) & \textbf{70\%} (0.7) \\
& Evolution Strategy & Dynamic Pruning ($0.3 \to 0.7$) \\
\midrule
\textbf{Training Config} & Total Rounds ($P_3$) & \textbf{15 Rounds}\\
& Local Epochs ($E$) & \textbf{1 Epoch} \\
& Fine-Tuning ($P_4$) & \textbf{0 Epochs} \\
& Pruning Frequency & Every 5 Rounds \\
\bottomrule
\end{tabular}%
}
\end{table}
\begin{table}[h]
\centering
\caption{Configuration for Fine-Tuning Ablation}
\label{tab:finetune_config}
\resizebox{\columnwidth}{!}{%
\renewcommand{\arraystretch}{1.2}
\begin{tabular}{l l c}
\toprule
\textbf{Component} & \textbf{Parameter} & \textbf{Value} \\
\midrule
\textbf{Dataset} & Scenario & WISDM (Standard) \\
 & Partitioning & Natural User Split \\
\midrule
\textbf{Phase 1, 2 and 3} & Rounds ($P_1, P_2, P_3$) & (0, 0, 50) \\
 & Sparsity Target & 70\% \\
 & Clients per Round & 10 \\
\midrule
\textbf{Phase 4: Personalization} & Rounds ($P_4$) & \{1, 5, 10, 15, 20, 25\} \\
 & Optimizer & Adam ($lr=1e^{-3}$) \\
\midrule
\textbf{Methods Compared} & CA-AFP & $K=3$, $S=70\%$\\
 & Global-FT (variant of CA-AFP) & \textbf{$K=1$} (No Clustering), $S=70\%$ \\
 & FedSNIP & $K=1$, $S=70\%$ \\
 & EfficientFL & $K=1$, $S=70\%$ \\
 & FedCHAR & $K=3$, $S=0\%$ \\
 & ClusterFL & $K=3$, $S=0\%$\\
\bottomrule
\end{tabular}%
}
\end{table}

\begin{table}[h]
\centering
\caption{Configuration for Non-IID Experiment}
\label{tab:noniid_setup}
\small
\renewcommand{\arraystretch}{1.1} 
\setlength{\tabcolsep}{5pt} 
\begin{tabular}{l l}
\hline
\textbf{Parameter} & \textbf{Configuration} \\
\hline
\textbf{Dataset} & WISDM (Activity Recognition) \\
\textbf{Total Clients} & 36 (Heterogeneous) \\
\textbf{Global Rounds} & 50 \\
\textbf{Target Sparsity} & 70\% ($0.7$) \\
\textbf{Runs per Scenario} & 5 Independent Seeds \\
\hline
\multicolumn{2}{c}{\textbf{Non-IID Scenarios}} \\
\hline
\textit{Extreme Non-IID} & Clients restricted to \textbf{1 class} only \\
\textit{Moderate Non-IID} & Clients restricted to \textbf{2 classes} only \\
\textit{Mild Non-IID} & Clients restricted to \textbf{3 classes} only \\
\hline
\end{tabular}
\end{table}
\clearpage

\end{document}